\documentclass[preprint,12pt]{elsarticle}

\usepackage{amssymb}
\usepackage{amsthm}
\usepackage{enumerate}
\newtheorem{thm}{Theorem}

\newtheorem{defn}{Definition}
\newtheorem{prop}{Proposition}
\newtheorem{exmp}{Example}

\newproof{pf}{Proof}
\newproof{ack}{Acknowledgements}

\begin{document}

\begin{frontmatter}



\title{Possibility neutrosophic soft sets with applications in decision making and similarity measure}


\author
{Faruk Karaaslan} \ead{fkaraaslan@karatekin.edu.tr}
\address{Department of Mathematics, Faculty of Sciences, \c{C}ank{\i}r{\i} Karatekin University, 18100, \c{C}ank{\i}r{\i},
Turkey}
\begin{abstract}In this paper, concept of possibility neutrosophic soft set and
its operations are defined, and their properties  are studied. An
application of this theory in decision making is investigated. Also
a similarity measure of two  possibility neutrosophic soft sets is
introduced  and discussed. Finally an application of this similarity
measure in personal selection for a firm.

\end{abstract}
\begin{keyword} Soft set, neutrosophic soft set, possibility
neutrosophic soft set, similarity measure, decision making
\end{keyword}

\end{frontmatter}


\section{Introduction}

In this physical world problems in engineering, medical sciences,
economics and social sciences the information involved are
uncertainty in nature. To cope with these problems, researchers
proposed some theories such as the theory of fuzzy set
\cite{zadeh-1965}, the theory of intuitionistic fuzzy set
\cite{atanassov-1986}, the theory of rough set \cite{paw-82}, the
theory of vague set \cite{gau-1993}. In 1999, Molodtsov
\cite{molodtsov-1999} initiated the theory of soft sets as a new
mathematical  tool for dealing with uncertainties as different from
these theories. A wide range of applications of soft sets have been
developed in many different fields, including the smoothness of
functions, game theory, operations research, Riemann integration,
Perron integration, probability theory and measurement theory. Maji
et al. \cite{maji-2002,maj-03sst}  applied soft set theory to
decision making problem and in 2003, they  introduced some new
operations of soft sets. After Maji's work, works on soft set theory
and its applications have been progressing rapidly. see
\cite{ali-09osnop,aktas-2007,cagman-2010,cagman-2014,feng-2010,feng-2012,feng-2013,neo-2011,pei-2005,sezgin-2011,xia-2012,yang-08,zhu-2013}.

Neutrosophy has been introduced by Smarandache \cite{smarandache-2005a,smarandache-2005b}
as a new branch of philosophy and generalization of fuzzy logic,
intuitionistic fuzzy logic, paraconsistent logic.
Fuzzy sets and  intuitionistic fuzzy
sets  are characterized by membership functions, membership and
non-membership functions, respectively.  In some real life problems
for proper  description of  an object in uncertain and ambiguous
environment, we need to handle the indeterminate and incomplete
information. But fuzzy sets and  intuitionistic fuzzy sets don't
handle the indeterminant and inconsistent  information. Thus
neutrosophic set is defined by Samarandache \cite{smarandache-2005b}, as a new mathematical tool for dealing
with problems involving incomplete, indeterminacy, inconsistent knowledge.

Maji \cite{maji-2013} introduced  concept of  neutrosophic soft set
and some operations of neutrosophic soft sets. Karaaslan
\cite{karaaslan-2014} redefined concept and operations of
neutrosophic soft sets as different from Maji's neutrosophic soft
set definition and operations. Recently, the properties and
applications on the neutrosophic soft sets have been studied
increasingly
\cite{bro-13gns,bro-13ins,bro-2014,deli-14ivnss,deli-14nsm,sahin-2014}.

Alkhazaleh et al \cite{alkhazaleh-2011} were  firstly introduced
concept of possibility fuzzy soft set and its operations.
They gave applications  of this theory in solving  a decision
making problem and they also introduced a similarity measure of two possibility fuzzy soft sets and discussed
their application in a medical diagnosis problem. In 2012, Bashir et al.
\cite{bashir-2012} introduced  concept of possibility intuitionistic fuzzy soft set and
its operations and discussed similarity measure of two possibility intuitionistic
fuzzy sets. They also gave an application of this similarity measure.

This paper is organized as follow: in Section 2, some basic
definitions and operations are given regarding neutrosophic soft set
required in this paper. In Section 3, possibility neutrosophic soft
set is defined as a generalization of possibility fuzzy soft set and
possibility intuitionistic fuzzy soft set introduced by Alkhazaleh
\cite{alkhazaleh-2011} and Bashir \cite{bashir-2012}, respectively.
In Section 4, a decision making method is given using the
possibility neutrosophic soft sets. In Section 5, similarity measure
is introduced between two possibility neutrosophic soft sets and in
Section 6, an application related personal selection for a firm is
given as regarding  this similarity measure method.

\section{Preliminary}

In this paper, we recall some definitions, operation  and their properties related to
the neutrosophic soft set \cite{karaaslan-2014,maji-2013} required in
this paper.

\begin{defn}\label{neutrosophicsoftset}
A neutrosophic soft set (or namely \emph{ns}-set) $f$ over $U$ is a
neutrosophic set valued function from $E$ to $\mathcal{N}(U)$.  It
can be written as
$$
f=\Big\{\big(e,\{\langle u,t_{f(e)}(u),i_{f(e)}(u),
f_{f(e)}(u)\rangle:u\in U\}\big): e\in E\Big\}
$$
where, $\mathcal{N}(U)$ denotes set of  all neutrosophic sets over
$U$. Note that if $f(e)=\big\{\langle u,0,1,1\rangle: u\in U\big\}$,
the element $(e, f(e))$ is not appeared in the neutrosophic soft set
$f$. Set of all \emph{ns}-sets over $U$ is denoted by $\mathbb{NS}$.
\end{defn}
\begin{defn}\cite{karaaslan-2014}\label{neutrosophicssubset} Let $f,g\in \mathbb{NS}$. $f$ is said to
be neutrosophic soft subset of $g$, if $t_{f(e)}(u)\leq
t_{g(e)}(u)$, $ i_{f(e)}(u)\geq i_{g(e)}(u)$ $f_{f(e)}(u) \geq
f_{g(e)}(u)$, $\forall e\in E$, $\forall u\in U$. We denote it by $f
\sqsubseteq g$. $f$ is said to be neutrosophic soft super set of $g$
if $g$ is a neutrosophic soft subset of $f$. We denote it by
$f\sqsupseteq g$.

If $f$ is neutrosophic soft subset of $g$ and $g$ is neutrosophic
soft subset of $f$. We denote it $f=g$
\end{defn}
\begin{defn}\cite{karaaslan-2014} \label{ns-empty} Let $f\in \mathbb{NS}$. If $t_{f(e)}(u)=0$
and $i_{f(e)}(u)=f_{f(e)}(u)=1$ for all $e\in E$ and for all $u\in
U$, then $f$ is called  null \emph{ns}-set and denoted by
$\tilde\Phi$.
\end{defn}
\begin{defn}\cite{karaaslan-2014}\label{universal} Let $f\in \mathbb{NS}$. If $t_{f(e)}(u)=1$ and
$i_{f(e)}(u)=f_{f(e)}(u)=0$ for all $e\in E$ and for all $u\in U$,
then $f$ is called  universal \emph{ns}-set and denoted by $\tilde
U$.
\end{defn}
\begin{defn}\cite{karaaslan-2014}\label{uni-int} Let $f,g\in \mathbb{NS}$. Then union and intersection of
ns-sets $f$ and $g$ denoted by $f\sqcup g$ and $f\sqcap g$
respectively,  are defined by as follow
\begin{eqnarray*}
f\sqcup g & = & \Big\{\big(e,\{\langle u,t_{f(e)}(u)\vee
t_{g(e)}(u),
i_{f(e)}(u)\wedge i_{g(e)}(u),\\
&&f_{f(e)}(u)\wedge f_{g(e)}(u)\rangle:u\in U\}\big):e\in E\Big\}.
\end{eqnarray*}
and \emph{ns}-intersection of $f$ and $g$ is defined as
\begin{eqnarray*}
f\sqcap g & = & \Big\{\big(e,\{\langle u,t_{f(e)}(u)\wedge
t_{g(e)}(u),
i_{f(e)}(u)\vee i_{g(e)}(u),\\
&&f_{f(e)}(u)\vee f_{g(e)}(u)\rangle: u\in U\}\big):e\in E\Big\}.
\end{eqnarray*}
\end{defn}
\begin{defn}\cite{karaaslan-2014}\label{ns-complement} Let $f,g\in \mathbb{NS}$. Then complement of ns-set $f$,
 denoted by $f^{\tilde c}$, is defined as follow
$$
f^{\tilde c}=\Big\{\big(e,\{\langle
u,f_{f(e)}(u),1-i_{f(e)}(u),t_{f(e)}(u)\rangle :u\in U\}\big):e\in
E\Big\}.
$$
\end{defn}
\begin{prop}\cite{karaaslan-2014} Let $f,g,h\in\mathbb{NS}$. Then,
\begin{enumerate}[\it i.]
\item $\tilde \Phi\sqsubseteq f$
\item $f\sqsubseteq \tilde U$
\item $f\sqsubseteq f $
\item $f\sqsubseteq g $ and $g\sqsubseteq h \Rightarrow$ $f\sqsubseteq h$
\end{enumerate}
\end{prop}
\begin{prop}\cite{karaaslan-2014} Let $f\in\mathbb{NS}$. Then
\begin{enumerate}[\it i.]
\item $\tilde \Phi^{\tilde c}=\tilde U$
\item $\tilde U^{\tilde c}=\tilde\Phi $
\item $(f^{\tilde c})^{\tilde c}=f$.
\end{enumerate}
\end{prop}
\begin{prop}\cite{karaaslan-2014}
Let $f,g,h\in\mathbb{NS}$. Then,
\begin{enumerate}[\it i.]
\item $f\sqcap f=f$ and $f\sqcup f=f$
\item $f\sqcap g=g\sqcap f$ and $f\sqcup g=g\sqcup f$
\item $f\sqcap\tilde\Phi=\tilde\Phi$ and $f\sqcap\tilde U=f$
\item $f\sqcup\tilde\Phi=f$ and $f\sqcup\tilde U=\tilde U$
\item $f\sqcap(g\sqcap h)=(f\sqcap g)\sqcap h$ and $f\sqcup(g\sqcup h)=(f\sqcup g)\sqcup h$
\item $f\sqcap(g\sqcup h)=(f\sqcap g)\sqcup(f\sqcap h)$ and
$f\sqcup(g\sqcap h)=(f\sqcup g)\sqcap(f\sqcup h).$
\end{enumerate}
\end{prop}
\proof The proof is clear from definition and operations of
neutrosophic soft sets.
\begin{thm}\cite{karaaslan-2014} \label{t-ns-demorgan}
Let $f,g\in\mathbb{NS}$. Then, De Morgan's law is valid.
\begin{enumerate}[\it i.]
\item $(f\sqcup g)^{\tilde c}=f^{\tilde c}\sqcap g^{\tilde c}$
\item $(f\sqcup g)^{\tilde c}=f^{\tilde c}\sqcap g^{\tilde c}$
\end{enumerate}
\end{thm}

\begin{defn}\cite{karaaslan-2014} \label{or} Let $f,g\in \mathbb{NS}$. Then  'OR' product  of
ns-sets $f$ and $g$ denoted by $f\wedge g$, is defined  as follow
\begin{eqnarray*}
f\bigvee g & = & \Big\{\big((e,e'),\{\langle u,t_{f(e)}(u)\vee
t_{g(e)}(u),
i_{f(e)}(u)\wedge i_{g(e)}(u),\\
&&f_{f(e)}(u)\wedge f_{g(e)}(u)\rangle:u\in U\}\big):(e,e')\in
E\times E\Big\}.
\end{eqnarray*}
\end{defn}
\begin{defn}\cite{karaaslan-2014} \label{and} Let $f,g\in \mathbb{NS}$. Then  'AND' product  of
ns-sets $f$ and $g$ denoted by $f\vee g$, is defined  as follow
\begin{eqnarray*}
f\bigwedge g & = & \Big\{\big((e,e'),\{\langle u,t_{f(e)}(x)\wedge
t_{g(e)}(u),
i_{f(e)}(u)\vee i_{g(e)}(u),\\
&&f_{f(e)}(u)\vee f_{g(e)}(u)\rangle:u\in U\}\big):(e,e')\in E\times
E\Big\}.
\end{eqnarray*}
\end{defn}

\begin{prop}\cite{karaaslan-2014} Let $f,g\in \mathbb{NS}$. Then,
\begin{enumerate}
\item $(f\bigvee g)^{\tilde c}=f^{\tilde c}\bigwedge g^{\tilde c}$
\item $(f\bigwedge g)^{\tilde c}=f^{\tilde c}\bigvee g^{\tilde c}$
\end{enumerate}
\end{prop}
\begin{pf} The proof is clear from Definition \ref{or} and \ref{and}.
\end{pf}

\begin{defn}\cite{alkhazaleh-2011} Let $U=\{u_1,u_2,...,u_n\}$ be the universal set of elements and $E=\{e_1,e_2,...,e_m\}$ be the universal set of parameters.
The pair $(U,E)$ will be called a soft universe. Let $F:E\to I^U$ and $\mu$ be a fuzzy subset of $E$, that is $\mu:E\to I^U$, where $I^U$ is
the collection of all fuzzy subsets of $U$. Let $F_{\mu}:E\to I^U\times I^U$ be a function defined  as follows:
$$
F_{\mu}(e)=(F(e)(u),\mu(e)(u)), \forall u\in U.
$$
Then $F_{\mu}$ is called a possibility fuzzy soft set (PFSS in short) over the soft universe $(U,E)$.
 For each parameter $e_i$, $F_{\mu}(e_i)=(F(e_i)(u),\mu(e_i)(u))$ indicates not only the degree of belongingness of the elements  of $U$
 in $F(e_i)$, but also the degree of possibility of belongingness of the elements of $U$ in $F(e_i)$, which is represented by $\mu(e_i)$.
\end{defn}

\begin{defn}\cite{bashir-2012} Let $U=\{u_1,u_2,...,u_n\}$ be the universal set of elements and $E=\{e_1,e_2,...,e_m\}$ be the universal set of parameters.
The pair $(U,E)$ will be called a soft universe. Let $F:E\to (I\times I)^U\times I^U$  where $(I\times I)^U$ is  the collection of all intuitionistic fuzzy subsets of $U$ and $I^U$ is the collection of all intuitionistic fuzzy subsets of $U$. Let $p$ be a fuzzy subset of $E$, that is, $p:E\to I^U$ and let
$F_p:E\to (I\times I)^U\times I^U$ be a function defined as follows:

$$
F_{p}(e)=(F(e)(u),p(e)(u)), F(e)(u)=(\mu(u),\nu(u))\forall u\in U.
$$

Then $F_{p}$ is called a possibility intuitionistic fuzzy soft set (PIFSS in short) over the soft universe $(U,E)$.
 For each parameter $e_i$, $F_{p}(e_i)=(F(e_i)(u),p(e_i)(u))$ indicates not only the degree of belongingness of the elements  of $U$
 in $F(e_i)$, but also the degree of possibility of belongingness of the elements of $U$ in $F(e_i)$, which is represented by $p(e_i)$.
\end{defn}

\section{Possibility neutrosophic soft sets}
In this section, we introduced the concept of possibility
neutrosophic soft set, possibility neutrosophic soft subset,
possibility neutrosophic soft null set, possibility neutrosophic
soft universal set and possibility neutrosophic soft set operations
such as union, intersection and complement.

Throughout paper $U$ is an initial universe, $E$ is a set of parameters and $\Lambda$ is an index set.

\begin{defn} Let $U$ be an initial universe,
$E$ be  a parameter set, $\mathcal{N(U)}$ be the collection of all
neutrosophic sets of $U$ and $I^U$ is collection of all fuzzy subset
of $U$. A possibility neutrosophic soft set $(PNS$-set$)$ $f_{\mu}$
over $U$ is defined by the set of ordered pairs

$$
f_{\mu}=\big\{\big(e_i,\{(\frac{u_j}{f(e_i)(u_j)},\mu(e_i)(u_j)):u_j\in
U\}\big): e_i\in E\big\}
$$

where, $i,j\in \Lambda$,  $f$ is a mapping given by $f:E\to \mathcal{N(U)}$ and $\mu(e_i)$
is a fuzzy set such that $\mu:E\to I^U$. Here, $\tilde f_{\mu}$ is a
mapping defined by $f_{\mu}:E\to \mathcal{N(U)}\times I^U$.

For each parameter  $e_i\in E$, $f(e_i)=\big\{\langle u_j,
t_{f(e_i)}(u_j),i_{f(e_i)}(u_j),f_{f(e_i)}(u_j)\rangle:u_j\in
U\big\}$ indicates neutrosophic value set of parameter $e_i$ and
where $t,i,f: U\to [0,1]$ are the membership functions of truth,
indeterminacy and falsity respectively of the element $u_j\in U$.
For each $u_j\in U$ and $e_i\in E$, $0\leq
t_{f(e_i)}(u_j)+i_{f(e_i)}(u_j)+f_{f(e_i)}(u_j)\leq 3$. Also
$\mu(e_i)$, degrees of possibility of belongingness of elements of
$U$ in $f(e_i)$. So we can write \footnotesize
$$f_{\mu}(e_i)=\Big\{\Big(\frac{u_1}{f(e_i)(u_1)},\mu(e_i)(u_1)\Big),\Big(\frac{u_2}{f(e_i)(u_2)},\mu(e_i)(u_2)\Big),...,\Big(\frac{u_n}{f(e_i)(u_n)},\mu(e_i)(u_n)\Big)\Big\}$$
\end{defn}

From now on, we will show set of all possibility neutrosophic soft
sets over $U$ with $\mathcal{PS}(U,E)$ such that $E$ is parameter
set.

\begin{exmp}\label{pnss-ex}  Let $U=\{u_1,u_2,u_3\}$ be a set of three cars. Let
$E=\{e_1,e_2,e_3\}$ be a set of qualities where $e_1=$cheap,
$e_2=$equipment, $e_3=$fuel consumption  and let $\mu:E\to I^{U}$.
We define a function $f_{\mu}: E\to \mathcal{N(U)}\times I^{U}$ as
follows:\\

 $f_{\mu}=\left\{
\begin{array}{c}
f_{\mu}(e_1)=\Big\{\Big(\frac{u_1}{(0.5,0.2,0.6)},
0.8\Big),\Big(\frac{u_2}{(0.7,0.3,0.5)},
0.4\Big),\Big(\frac{u_3}{(0.4,0.5,0.8)},
0.7\Big)\Big\} \\
f_{\mu}(e_2)=\Big\{\Big(\frac{u_1}{(0.8,0.4,0.5)},
0.6\Big),\Big(\frac{u_2}{(0.5,0.7,0.2)},
0.8\Big),\Big(\frac{u_3}{(0.7,0.3,0.9)},
0.4\Big)\Big\}\\
f_{\mu}(e_3)=\Big\{\Big(\frac{u_1}{(0.6,0.7,0.5)},
0.2\Big),\Big(\frac{u_2}{(0.5,0.3,0.7)},
0.6\Big),\Big(\frac{u_3}{(0.6,0.5,0.4)}, 0.5\Big)\Big\}
\end{array}\right\}$\\

also we can define a function $g_{\nu}: E\to \mathcal{N(U)}\times
I^{U}$ as follows:\\

$g_{\nu}=\left\{\begin{array}{c}
g_{\nu}(e_1)=\Big\{\Big(\frac{u_1}{(0.6,0.3,0.8)},
0.4\Big),\Big(\frac{u_2}{(0.6,0.5,0.5)},
0.7),\Big(\frac{u_3}{(0.2,0.6,0.4)},
0.8\Big)\Big\} \\
g_{\nu}(e_2)=\Big\{\Big(\frac{u_1}{(0.5,0.4,0.3)},
0.3\Big),\Big(\frac{u_2}{(0.4,0.6,0.5)},
0.6\Big),\Big(\frac{u_3}{(0.7,0.2,0.5)},
0.8\Big)\Big\}\\
g_{\nu}(e_3)=\Big\{\Big(\frac{u_1}{(0.7,0.5,0.3)},
0.8\Big),\Big(\frac{u_2}{(0.4,0.4,0.6)},
0.5\Big),\Big(\frac{u_3}{(0.8,0.5,0.3)}, 0.6\Big)\Big\}
\end{array}\right\}$

\end{exmp}
For the purpose of storing a possibility neutrosophic soft set in a
computer, we can use matrix notation of possibility neutrosophic
soft set $f_{\mu}$. For example, matrix notation of possibility
neutrosophic soft set $f_{\mu}$ can be written as follows: for
$m,n\in \Lambda$,
$$f_{\mu}=\left(
  \begin{array}{ccc}
    (\langle0.5,0.2,0.6\rangle,0.8) & (\langle0.7,0.3,0.5\rangle,0.4) & (\langle0.4,0.5,0.8\rangle,0.7) \\
    (\langle0.8,0.4,0.5\rangle,0.6) & (\langle0.5,0.7,0.2\rangle,0.8) & (\langle0.7,0.3,0.9\rangle,0.4) \\
    (\langle0.6,0.7,0.5\rangle,0.2) & (\langle0.5,0.3,0.7\rangle, 0.6) & (\langle0.6,0.5,0.4\rangle, 0.5) \\
  \end{array}
\right)$$ where the $m-$th row vector shows $f(e_m)$ and $n-$th
column vector shows $u_n$.

\begin{defn} Let $f_{\mu}$, $g_{\nu}\in \mathcal{PS}(U,E)$. Then,  $f_{\mu}$
is said to be a possibility neutrosophic soft subset $(PNS$-subset$)$
of $g_{\nu}$, and denoted by $f_{\mu}\subseteq g_{\nu}$, if
\begin{enumerate}[{i.}]
    \item $\mu(e)$ is a fuzzy subset of $\nu(e)$, for all $e\in E$
    \item $f$ is a neutrosophic subset of $g$,
\end{enumerate}
\end{defn}

\begin{exmp} Let $U=\{u_1,u_2,u_3\}$ be a set of tree houses, and
let $E=\{e_1,e_2,e_3\}$ be a set of parameters where $e_1=$modern,
$e_2=$big and $e_3=$cheap. Let $f_{\mu}$ be a $PNS$-set defined as
follows:\\

$f_{\mu}=\left\{ \begin{array}{c}
f_{\mu}(e_1)=\Big\{\Big(\frac{u_1}{(0.5,0.2,0.6)},0.8\Big),\Big(\frac{u_2}{(0.7,0.3,0.5)},0.4\Big), \Big(\frac{u_3}{(0.4,0.5,0.9)},0.7\Big)\Big\} \\
f_{\mu}(e_2)=\Big\{\Big(\frac{u_1}{(0.8,0.4,0.5)},0.6\Big),\Big(\frac{u_2}{(0.5,0.7,0.2)},0.8\Big), \Big(\frac{u_3}{(0.7,0.3,0.9)},0.4\Big)\Big\} \\
f_{\mu}(e_3)=\Big\{\Big(\frac{u_1}{(0.6,0.7,0.5)},0.2\Big), \Big(\frac{u_2}{(0.5,0.3,0.8)}, 0.6\Big),\Big(\frac{u_3}{(0.6,0.5,0.4)},0.5\Big) \Big\}\\
\end{array}\right\}$\\

 $g_{\nu}:E\to\mathcal{N}(U)\times I^U$ be another $PNS$-set
defined as follows:\\

$g_{\nu}=\left\{ \begin{array}{c}
g_{\nu}(e_1)=\Big\{\Big(\frac{u_1}{(0.6,0.1,0.5)},0.9\Big),\Big(\frac{u_2}{(0.8,0.2,0.3)},0.6\Big), \Big(\frac{u_3}{(0.7,0.5,0.8)},0.8\Big)\Big\} \\
g_{\nu}(e_2)=\Big\{\Big(\frac{u_1}{(0.9,0.2,0.4)},0.7\Big),\Big(\frac{u_2}{(0.9,0.5,0.1)},0.9\Big), \Big(\frac{u_3}{(0.8,0.1,0.9)},0.5\Big)\Big\} \\
g_{\nu}(e_3)=\Big\{\Big(\frac{u_1}{(0.6,0.5,0.4)},0.4\Big), \Big(\frac{u_2}{(0.7,0.1,0.7)}, 0.9\Big),\Big(\frac{u_3}{(0.8,0.2,0.4)},0.7\Big) \Big\}\\
\end{array}\right\}$\\
\end{exmp}
it is clear that $f_{\mu}$ is $PNS-subset$ of $g_{\nu}$

\begin{defn}\label{pns-subset}
Let $f_{\mu}, g_{\nu}\in \mathcal{PN}(U,E)$. Then, $f_{\mu}$ and $g_{\nu}$
are called possibility neutrosophic soft equal set and denoted by
$f_{\mu}=g_{\nu}$, if $f_{\mu}\subseteq g_{\nu}$  and
$f_{\mu}\supseteq g_{\nu}$.
\end{defn}

\begin{defn}\label{pns-null}
Let $f_{\mu}\in  \mathcal{PN}(U,E)$. Then,  $f_{\mu}$ is said to be
 possibility  neutrosophic soft null set denoted by $\phi_{\mu}$, if $\forall e\in E$, $\phi_{\mu}:E \to \mathcal{N(U)}\times
 I^U$ such that $\phi_{\mu}(e)=\{(\frac{u}{\phi(e)(u)},\mu(e)(u)): u\in U\}$, where $\phi(e)=\{\langle
 u,0,1,1\rangle: u\in U\}$ and $\mu(e)=\{(u,0): u\in U\})$.
\end{defn}

\begin{defn}\label{pns-universal}
Let $f_{\mu}\in  \mathcal{PN}(U,E)$. Then,  $f_{\mu}$ is said to be
possibility  neutrosophic soft universal set denoted by $U_{\mu}$, if $\forall e\in E$, $U_{\mu}:E \to \mathcal{N(U)}\times
 I^U$ such that $U_{\mu}(e)=\{(\frac{u}{U(e)(u)},\mu(e)(u)): u\in U\}$, where $U(e)=\{\langle
 u,1,0,0\rangle: u\in U\}$ and $\mu(e)=\{(u,1): u\in U\})$.
\end{defn}

\begin{prop}Let $f_{\mu},g_{\nu}$ and $h_{\delta}\in \mathcal{PN}(U,E)$. Then,
\begin{enumerate}[\it i.]
\item $\phi_{\mu}\subseteq f_{\mu}$
\item $f_{\mu}\subseteq U_{\mu}$
\item $f_{\mu}\subseteq f_{\mu} $
\item $f_{\mu}\subseteq g_{\nu} $ and $g_{\nu}\subseteq h_{\delta} \Rightarrow$ $f_{\mu}\subseteq h_{\delta}$
\end{enumerate}
\end{prop}
\begin{pf}
It is clear from Definition \ref{pns-subset}, \ref{pns-null} and \ref{pns-universal}.
\end{pf}

\begin{defn} Let $f_{\mu}\in \mathcal{PN}(U)$, where $f_{\mu}(e_i)=\{(f(e_i)(u_j),\mu(e_i)(u_j)):e_i\in E, u_j\in U\}$
and $f(e_i)=\big\{\langle u,
t_{f(e_i)}(u_j),i_{f(e_i)}(u_j),f_{f(e_i)}(u_j)\rangle\big\}$ for
all $e_i\in E$, $u\in U$. Then for $e_i \in E$ and $u_j\in U$,

\begin{enumerate}
    \item $f_{\mu}^t$ is said to be truth-membership part of $f_{\mu}$,\\
    $f_{\mu}^t=\{(f^t_{ij}(e_i),\mu_{ij}(e_i))\}$ and $f^t_{ij}(e_i)=\{( u_j,
    t_{f(e_i)}(u_j))\}$, $\mu_{ij}(e_i)=\{(u_j,\mu(e_i)(u_j))\}$
    \item $f_{\mu}^i$ is said to be indeterminacy-membership part of $f_{\mu}$,\\
    $f_{\mu}^i=\{(f^i_{ij}(e_i),\mu_{ij}(e_i))\}$ and $f^i_{ij}(e_i)=\{( u_j,
    i_{f(e_i)}(u_j))\}$, $\mu_{ij}(e_i)=\{(u_j,\mu(e_i)(u_j))\}$
    \item $f_{\mu}^f$ is said to be truth-membership part of $f_{\mu}$,\\
    $f_{\mu}^f=\{(f^f_{ij}(e_i),\mu_{ij}(e_i))\}$ and $f^f_{ij}(e_i)=\{( u_j,
    f_{f(e_i)}(u_j))\}$, $\mu_{ij}(e_i)=\{(u_j,\mu(e_i)(u_j))\}$
\end{enumerate}
We can write a possibility neutrosophic soft set in form
$f_{\mu}=(f_{\mu}^t,f_{\mu}^i,f_{\mu}^f)$.
\end{defn}
If  considered the possibility neutrosophic soft set $f_{\mu}$ in
Example \ref{pnss-ex}, $f_{\mu}$ can be expressed in matrix form as
follow:

$$f_{\mu}^t=\left(
  \begin{array}{ccc}
  (0.5,0.8) & (0.7,0.4) &(0.4,0.7)   \\
    (0.8,0.6)& (0.5,0.8) & (0.7,0.4)  \\
    (0.6,0.2) & (0.5,0.6) & (0.6,0.5)  \\
     \end{array}
\right)$$
$$f_{\mu}^i=\left(
  \begin{array}{ccc}
   (0.2,0.8) & (0.3,0.4) & (0.5,0.7)    \\
   (0.4,0.6) & (0.7,0.8) & (0.3,0.4)  \\
   (0.7,0.2) & (0.3,0.6) & (0.5,0.5) \\
     \end{array}
\right)$$
$$f_{\mu}^f=\left(
  \begin{array}{ccc}
      (0.6,0.8) & (0.5,0.4)& (0.8,0.7) \\
      (0.5,0.6) & (0.2,0.8) & (0.9,0.4)\\
     (0.5,0.2) &(0.7,0.6) & (0.4,0.5)\\
     \end{array}
\right)
$$

\begin{defn}\cite{schweirer-1960}
A binary operation $\otimes:[0,1]\times [0,1] \to [0,1]$ is
continuous $t-$norm if $\otimes$ satisfies the following conditions
\begin{enumerate}[\it i.]
    \item $\otimes$ is commutative and associative,
    \item $\otimes$ is continuous,
    \item $a\otimes 1=a$, $\forall a\in [0,1]$,
    \item $a\otimes b\leq c\otimes d$ whenever $a\leq c, b\leq
    d$ and $a,b,c,d\in [0,1]$.
\end{enumerate}
\end{defn}

\begin{defn}\cite{schweirer-1960}
A binary operation $\oplus:[0,1]\times [0,1] \to [0,1]$ is
continuous $t-$conorm (s-norm) if $\oplus$ satisfies the following
conditions
\begin{enumerate}[\it i.]
    \item $\oplus$ is commutative and associative,
    \item $\oplus$ is continuous,
    \item $a\oplus 0=a$, $\forall a\in [0,1]$,
    \item $a \oplus b\leq c\oplus d$ whenever $a\leq c, b\leq
    d$ and $a,b,c,d\in [0,1]$.
\end{enumerate}
\end{defn}

\begin{defn}
Let $I^3=[0,1]\times [0,1]\times [0,1]$ and $N(I^3)=\{(a,b,c):
a,b,c\in [0,1]\}$. Then $(N(I^3),\oplus, \otimes)$ be a lattices
together with partial ordered relation $\preceq$, where order
relation $\preceq$ on $N(I^3)$ can be defined by for  $(a,b,c),
(d,e,f)\in N(I^3)$
$$
(a,b,c)\preceq (e,f,g)\Leftrightarrow a\leq e, b\geq f, c\geq g
$$
\end{defn}

\begin{defn}A binary operation $$\tilde\otimes:\Big([0,1]\times [0,1]\times
[0,1]\Big)^2 \to [0,1]\times [0,1]\times
[0,1]$$ is
continuous $n-$norm  if $\tilde\otimes$ satisfies the following
conditions
\begin{enumerate}[\it i.]
    \item $\tilde\otimes$ is commutative and associative,
    \item $\tilde\otimes$ is continuous,
    \item $a\tilde\otimes 0=0$, $a\tilde\otimes1=a$, $\forall a\in [0,1]\times [0,1]\times
[0,1]$, $(1=(1,0,0))\,\,and \,(0=(0,1,1))$
    \item $a \tilde\otimes b\leq c\tilde\otimes d$ whenever $a\preceq c, b\preceq
    d$ and $a,b,c,d\in [0,1]\times [0,1]\times
[0,1]$.
\end{enumerate}
Here, $$a\tilde\otimes b=\tilde\otimes(\langle
t(a),i(a),f(a)\rangle,\langle t(b),i(b),f(b)\rangle)=\langle
t(a)\otimes t(b), i(a)\oplus i(b), f(a)\oplus f(b)$$
\end{defn}

\begin{defn}A binary operation $$\tilde\oplus:\Big([0,1]\times [0,1]\times
[0,1]\Big)^2 \to [0,1]\times [0,1]\times
[0,1]$$ is
continuous $n-$conorm  if $\tilde\oplus$ satisfies the following
conditions
\begin{enumerate}[\it i.]
    \item $\tilde\oplus$ is commutative and associative,
    \item $\tilde\oplus$ is continuous,
    \item $a\tilde\oplus 0=a$, $a\tilde\oplus1=1$, $\forall a\in [0,1]\times [0,1]\times
[0,1]$, $(1=(1,0,0))\,\,and \,(0=(0,1,1))$
    \item $a \tilde\oplus b\leq c\tilde\oplus d$ whenever $a\leq c, b\leq
    d$ and $a,b,c,d\in [0,1]\times [0,1]\times
[0,1]$.
\end{enumerate}
Here, $$a \tilde\oplus b=\tilde\oplus(\langle
t(a),i(a),f(a)\rangle,\langle t(b),i(b),f(b)\rangle)=\langle
t(a)\oplus t(b), i(a)\otimes i(b), f(a)\otimes f(b)$$
\end{defn}

\begin{defn}\label{pns-uni} Let $f_{\mu}, g_{\nu}\in \mathcal{PN}(U,E).$ The union
of two possibility neutrosophic soft sets $f_{\mu}$ and $ g_{\nu}$
over $U$, denoted by $f_{\mu}\cup g_{\nu}$ is defined by as follow:
\scriptsize
$$
f_{\mu}\cup g_{\nu}=\Big\{\Big(e_i,\Big\{\big(\frac{u_j}{(f^t_{ij}(e_i)\oplus
g^t_{ij}(e_i), f^i_{ij}(e_i)\otimes g^i_{ij}(e_i),
f^f_{ij}(e_i)\otimes g^f_{ij}(e_i))},
\mu_{ij}(e_i)\oplus\nu_{ij}(e_i)\big): u_j\in U\Big\}\Big): e_i\in E \Big\}
$$
\end{defn}

\begin{defn}\label{pns-int} Let $f_{\mu}, g_{\nu}\in \mathcal{PN}(U,E).$ The
intersection of two possibility neutrosophic soft sets $f_{\mu}$ and
$ g_{\nu}$ over $U$, denoted by $f_{\mu}\cap g_{\nu}$ is defined by
as follow: \scriptsize
$$
f_{\mu}\cap g_{\nu}=\Big\{\Big(e_i,\Big\{\big(\frac{u_j}{(f^t_{ij}(e_i)\otimes
g^t_{ij}(e_i), f^i_{ij}(e_i)\oplus g^i_{ij}(e_i),
f^f_{ij}(e_i)\oplus g^f_{ij}(e_i))},
\mu_{ij}(e_i)\otimes\nu_{ij}(e_i)\big): u_j\in U\Big\}\Big): e_i\in E \Big\}
$$
\end{defn}

\begin{exmp} Let us consider the possibility neutrosophic soft sets
$f_{\mu}$ and $g_{\nu}$ defined as in Example \ref{pnss-ex}. Let us
suppose that $t-$norm is defined by $a\otimes b=min\{a,b\}$ and the
$t-$conorm is defined by $a\oplus b=max\{a,b\}$ for $a,b\in [0,1]$.
Then,
\footnotesize

$f_{\mu}\cup g_{\nu}=\left\{\begin{array}{c} (f_{\mu}\cup
g_{\nu})(e_1)=\Big\{\Big(\frac{u_1}{(0.6,0.2,0.6)},0.8\Big),\Big(\frac{u_2}{(0.7,0.3,0.5)},0.7\Big),\Big(\frac{u_3}{(0.4,0.5,0.4)},0.8\Big)\Big\}\\
(f_{\mu}\cup g_{\nu})(e_2)=\Big\{\Big(\frac{u_1}{(0.8,0.4,0.3)},0.6\Big),\Big(\frac{u_2}{(0.5,0.6,0.2)},0.8\Big),\Big(\frac{u_3}{(0.7,0.2,0.5)},0.8\Big)\Big\}\\
(f_{\mu}\cup g_{\nu})(e_3)=\Big\{\Big(\frac{u_1}{(0.7,0.3,0.3)},0.8\Big),\Big(\frac{u_2}{(0.5,0.3,0.6)},0.6\Big),\Big(\frac{u_3}{(0.8,0.5,0.3)},0.6\Big)\Big\}\\
\end{array}\right\}$\\

and
\\

\footnotesize
 $f_{\mu}\cap g_{\nu}=\left\{\begin{array}{c} (f_{\mu}\cap
g_{\nu})(e_1)=\Big\{\Big(\frac{u_1}{(0.5,0.3,0.8)},0.4\Big),\Big(\frac{u_2}{(0.6,0.5,0.5)},0.4\Big),\Big(\frac{u_3}{(0.2,0.6,0.8)},0.7\Big)\Big\}\\
(f_{\mu}\cap
g_{\nu})(e_2)=\Big\{\Big(\frac{u_1}{(0.5,0.4,0.5)},0.3\Big),\Big(\frac{u_2}{(0.4,0.7,0.5)},0.6\Big),\Big(\frac{u_3}{(0.7,0.3,0.9)},0.4\Big)\Big\}\\
(f_{\mu}\cap
g_{\nu})(e_3)=\Big\{\Big(\frac{u_1}{(0.6,0.7,0.5)},0.2\Big),\Big(\frac{u_2}{(0.4,0.5,0.7)},0.5\Big),\Big(\frac{u_3}{(0.6,0.5,0.4)},0.5\Big)\Big\}\\
\end{array}\right\}$
\end{exmp}

\begin{prop}
Let $f_{\mu},g_{\nu},h_{\delta}\in\mathcal{PN}(U,E)$. Then,
\begin{enumerate}[\it i.]
\item $f_{\mu}\cap f_{\mu}=f_{\mu}$ and $f_{\mu}\cup f_{\mu}=f_{\mu}$
\item $f_{\mu}\cap g_{\nu}=g_{\nu}\cap f_{\mu}$ and $f_{\mu}\cup g_{\nu}=g_{\nu}\cup f_{\mu}$
\item $f_{\mu}\cap\phi_{\mu}=\phi_{\mu}$ and $f_{\mu}\cap U_{\mu}=f_{\mu}$
\item $f_{\mu}\cup\phi=f_{\mu}$ and $f_{\mu}\cup U_{\mu}= U_{\mu}$
\item $f_{\mu}\cap(g_{\nu}\cap h_{\delta})=(f_{\mu}\cap g_{\nu})\cap h_{\delta}$ and $f_{\mu}\cup(g_{\nu}\cup h_{\delta})=(f_{\mu}\cup g_{\nu})\cup h_{\delta}$
\item $f_{\mu}\cap(g_{\nu}\cup h_{\delta})=(f_{\mu}\cap g_{\nu})\cup(f_{\mu}\cap h_{\delta})$ and
$f_{\mu}\cup(g_{\nu}\cap h_{\delta})=(f_{\mu}\cup
g_{\nu})\cap(f_{\mu}\cup h_{\delta}).$
\end{enumerate}
\end{prop}
\begin{pf}
The proof can be obtained from Definitions \ref{pns-uni}. and
\ref{pns-int}.
\end{pf}

\begin{defn}\cite{fodor-1994,trillas-1979} A function $N:[0,1]\to [0,1]$ is called a negation if $N(0)=1$, $N(1)=0$
and $N$ is non-increasing $(x\leq y \Rightarrow N(x)\geq N(y))$. A
negation is called a strict negation if it is strictly decreasing
$(x<y \Rightarrow N(x)>N(y))$ and continuous. A strict negation is
said to be  a strong negation if it is also involutive, i.e.
$N(N(x))=x$
\end{defn}

\begin{defn}\cite{smarandache-2005a} A function $n_N:[0,1]\times [0,1]\times [0,1]\to [0,1]\times[0,1]\times[0,1]$
 is called a negation if $n_N(\overline{0})=\overline{1}$, $n_N(\overline{1})=\overline{0}$
and $n_N$ is non-increasing $(x\preceq y \Rightarrow n_N(x)\succeq
n_N(y))$. A negation is called a strict negation if it is strictly
decreasing $(x\prec y \Rightarrow N(x)\succ N(y))$ and continuous.
\end{defn}

\begin{defn}\label{pns-comp} Let $f_{\mu}\in \mathcal{PN}(U,E).$
Complement of possibility neutrosophic soft set $f_{\mu}$, denoted
by $f^c_{\mu}$ is defined as follow:

$$
f^c_{\mu}=\Big\{\Big(e,\big\{\big(\frac{u_j}{n_N(f(e_i))},
N(\mu_{ij}(e_i)(u_j))\Big): u_j\in U \big\}\Big): e\in E\Big\}
$$
where
$(n_N(f_{ij}(e_i))=(N(f^t_{ij}(e_i)),N(f^i_{ij}(e_i)),N(f^f_{ij}(e_i))$
for all $i,j\in \Lambda$
\end{defn}

\begin{exmp}
Let us consider the possibility neutrosophic soft set $f_{\mu}$
define in Example \ref{pnss-ex}. Suppose that the negation is
defined by $N(f^t_{ij}(e_i))=f^f_{ij}(e_i)$,
$N(f^f_{ij}(e_i))=f^t_{ij}(e_i)$, $N(f^i_{ij}(e_i))=1-f^i_{ij}(e_i)$
and $N(\mu_{ij}(e_i))=1-\mu_{ij}(e_i)$, respectively. Then,
$f^c_{\mu}$ is defined as follow:\\

$f^c_{\mu}=\left\{
\begin{array}{c}
f^c_{\mu}(e_1)=\Big\{\Big(\frac{u_1}{(0.6,0.8,0.5)},
0.2\Big),\Big(\frac{u_2}{(0.5,0.7,0.7)},
0.6\Big),\Big(\frac{u_3}{(0.8,0.5,0.4)},
0.3\Big)\Big\} \\
f^c_{\mu}(e_2)=\Big\{\Big(\frac{u_1}{(0.5,0.6,0.8)},
0.4\Big),\Big(\frac{u_2}{(0.2,0.3,0.5)},
0.2\Big),\Big(\frac{u_3}{(0.9,0.7,0.7)},
0.6\Big)\Big\}\\
f^c_{\mu}(e_3)=\Big\{\Big(\frac{u_1}{(0.5,0.3,0.6)},
0.8\Big),\Big(\frac{u_2}{(0.7,0.7,0.5)},
0.4\Big),\Big(\frac{u_3}{(0.4,0.5,0.6)}, 0.5\Big)\Big\}
\end{array}\right\}$\\

\end{exmp}

\begin{prop}Let $f_{\mu}\in\mathcal{PN}(U,E)$. Then,
\begin{enumerate}[\it i.]
\item $\phi^c_{\mu}= U_{\mu}$
\item $U^c_{\mu}=\phi_{\mu} $
\item $(f^c_{\mu})^{c}=f_{\mu}$.
\end{enumerate}
\begin{pf} It is clear from Definition \ref{pns-comp}
\end{pf}
\end{prop}

\begin{prop}
Let $f_{\mu},g_{\nu}\in\mathcal{PN}(U,E)$. Then, De Morgan's law is
valid.
\begin{enumerate}[\it i.]
\item $(f_{\mu}\cup g_{\nu})^{c}=f^c_{\mu}\cap g^c_{\nu}$
\item $(f_{\mu}\cap g_{\nu})^{c}=f^c_{\mu}\cup g^c_{\nu}$
\end{enumerate}
\end{prop}
\begin{pf}
\begin{enumerate}[\it i.]
\item Let $i,j\in \Lambda$
\tiny
\begin{eqnarray*}
(f_{\mu}\cup
g_{\nu})^{c}&=&\Big\{\Big(e_i,\big\{\big(\frac{u_j}{(f^t_{ij}(e_i)\oplus
g^t_{ij}(e_i), f^i_{ij}(e_i)\otimes g^i_{ij}(e_i),
f^f_{ij}(e_i)\otimes
g^f_{ij}(e_i))},\\
&&\mu_{ij}(e_i)\oplus\nu_{ij}(e_i)\big):
u_j\in U\big\}\Big): e_i\in E \Big\}^c\\
&=&\Big\{\Big(e_i,\big\{\big(\frac{u_j}{(f^f_{ij}(e_i)\otimes
g^f_{ij}(e_i), N(f^i_{ij}(e_i)\otimes g^i_{ij}(e_i)),
f^t_{ij}(e_i)\oplus
g^t_{ij}(e_i))},\\
&&N(\mu_{ij}(e_i)\oplus\nu_{ij}(e_i))\big):
u_j\in U\big\}\Big): e_i\in E \Big\}\\
&=&\Big\{\Big(e_i,\big\{\big(\frac{u_j}{(f^f_{ij}(e_i)\otimes
g^f_{ij}(e_i), N(f^i_{ij}(e_i))\oplus N(g^i_{ij}(e_i))),
f^t_{ij}(e_i)\oplus g^t_{ij}(e_i))},\\
&&N(\mu_{ij}(e_i))\otimes
N(\nu_{ij}(e_i))\big):
u_j\in U\big\}\Big): e_i\in E \Big\}\\
&=&\Big\{\Big(e_i,\big\{\big(\frac{u_j}{(f^f_{ij}(e_i),
N(f^i_{ij}(e_i)), f^t_{ij}(e_i)},N(\mu_{ij}(e_i))\big):
u_j\in U\big\}\Big): e_i\in E \Big\}\\
&\cap& \Big\{\Big(e_i,\big\{\big(\frac{u_j}{ g^f_{ij}(e_i),
N(g^i_{ij}(e_i)), g^t_{ij}(e_i))},  N(\nu_{ij}(e_i))\big):
u_j\in U\Big\}\Big): e_i\in E \Big\}\\
&=&\Big\{\Big(e_i,\big\{\big(\frac{u_j}{(f^t_{ij}(e_i),
N(f^i_{ij}(e_i)), f^f_{ij}(e_i)},\mu_{ij}(e_i)\big):
u_j\in U\big\}\Big): e_i\in E \Big\}^c\\
&\cap& \Big\{\Big(e_i,\big\{\big(\frac{u_j}{ g^t_{ij}(e_i),
g^i_{ij}(e_i), g^f_{ij}(e_i))}, \nu_{ij}(e_i)\big):
u_j\in U\big\}\Big): e_i\in E \Big\}^c\\
&=&f_{\mu}^c\cap g_{\nu}^c
\end{eqnarray*}
\normalsize
\item The proof can be made with similar way.
\end{enumerate}
\end{pf}

\begin{defn} Let $f_{\mu}$ and $g_{\nu}\in \mathcal{PN}(U,E)$.  Then  'AND' product  of
PNS-set $f_{\mu}$ and $g_{\nu}$ denoted by $f_{\mu}\wedge g_{\nu} $,
is defined as follow:

\small
\begin{eqnarray*}f_{\mu}\wedge g_{\nu}&=&\Big\{\Big((e_k,e_l), (f^t_{kj}(e_k)\wedge
g^t_{lj}(e_l),f^i_{kj}(e_k)\vee g^i_{lj}(e_l),f^f_{kj}(e_k)\vee
g^f_{lj}(e_l)),\\
&&\mu_{kj}(e_k)\wedge \nu_{lj}(e_l)\Big): (e_k,e_l)\in
E\times E, j,k,l \in \Lambda \Big\}\\
\end{eqnarray*}
\end{defn}

\begin{defn} Let $f_{\mu}$ and $g_{\nu}\in \mathcal{PN}(U,E)$.  Then  'OR' product  of
PNS-set $f_{\mu}$ and $g_{\nu}$ denoted by $f_{\mu}\vee g_{\nu} $,
is defined as follow:

\small
\begin{eqnarray*}f_{\mu}\vee g_{\nu}&=&\Big\{\Big((e_k,e_l),
(f^t_{kj}(e_k)\vee g^t_{lj}(e_l),f^i_{kj}(e_k)\wedge
g^i_{lj}(e_l),f^f_{kj}(e_k)\wedge
g^f_{lj}(e_l)),\\
&&\mu_{kj}(e_k)\vee \nu_{lj}(e_l)\Big): (e_k,e_l)\in
E\times E, j,k,l \in \Lambda \Big\}\\
\end{eqnarray*}
\end{defn}

\section{Decision making method}
In this section we will construct a decision making  method over the
possibility neutrosophic soft set. Firstly, we will define  some notions that
necessary  to construct algorithm of decision making method. And then
we will present an application of possibility neutrosophic soft set theory in a decision making problem.

\begin{defn}\label{weighedmatrices} Let $f_{\mu}\in  \mathcal{PN}(U,E)$, $f^t_{\mu},f^i_{\mu}$ and  $f^f_{\mu}$ be the truth,
indeterminacy and falsity matrices of $\wedge$-product matrix, respectively. Then, weighted matrices of $f^t_{\mu},f^i_{\mu}$ and  $f^f_{\mu}$
are denoted by $\wedge^t, \wedge^i$ and $\wedge^f$, respectively and computed by as follows:\\

\footnotesize
$\wedge^t(e_{ij},u_k)=t_{(f_{\mu}\wedge g_{\nu})(e_{ij})}(u_k)+
(\mu_{ik}(e_i)\wedge \nu_{jk}(e_j))-t_{(f_{\mu}\wedge
g_{\nu})(e_{ij})}(u_k)\times (\mu_{ik}(e_i)\wedge\nu_{jk}(e_j))$\\

$\wedge^i(e_{ij},u_k)=i_{(f_{\mu}\wedge g_{\nu})(e_{ij})}(u_k)\times
(\mu_{ik}(e_i)\wedge \nu_{jk}(e_j))$\\

$\wedge^f(e_{ij},u_k)=f_{(f_{\mu}\wedge g_{\nu})(e_{ij})}(u_k)\times
(\mu_{ik}(e_i)\wedge \nu_{jk}(e_j))$\\

\normalsize
for $i,j,k\in \Lambda$
\end{defn}

\begin{defn}\label{score} Let $f_{\mu}\in  \mathcal{PN}(U,E)$, $\wedge^t, \wedge^i$ and
$\wedge^f$ be the weighed matrices of  $f^t_{\mu},f^i_{\mu}$ and  $f^f_{\mu}$, respectively.
Then, for all $u_t\in U$ such that $t\in \Lambda$, scores of $u_t$ is in the
weighted matrices $\wedge^t, \wedge^i$ and $\wedge^f$ are
denoted by $s^t(u_k)$, $s^i(u_k)$ and $s^f(u_k)$, and computed by as follow, respectively

$$
s^t(u_t)=\sum_{i,j\in\Lambda}\delta^t_{ij}(u_t)
$$
$$
s^i(u_t)=\sum_{i,j\in\Lambda}\delta^i_{ij}(u_t)
$$
$$
s^f(u_t)=\sum_{i,j\in\Lambda}\delta^f_{ij}(u_t)
$$
here
$$\delta^t_{ij}(u_t)=\left\{
                       \begin{array}{ll}
                         \wedge^t(e_{ij},u_t), & \wedge^t(e_{ij},u_t)=max\{\wedge^t(e_{ij},u_k): u_k\in U\} \\
                         0, & otherwise
                       \end{array}
                     \right.
$$
$$\delta^i_{ij}(u_t)=\left\{
                       \begin{array}{ll}
                         \wedge^i(e_{ij},u_t), & \wedge^i(e_{ij},u_t)=max\{\wedge^i(e_{ij},u_k): u_k\in U\} \\
                         0, & otherwise
                       \end{array}
                     \right.
$$
$$\delta^f_{ij}(u_t)=\left\{
                       \begin{array}{ll}
                         \wedge^f(e_{ij},u_t), & \wedge^f(e_{ij},u_t)=max\{\wedge^f(e_{ij},u_k): u_k\in U\} \\
                         0, & otherwise
                       \end{array}
                     \right.
$$
\end{defn}
\begin{defn}
Let $s^t(u_t), s^i(u_t)$ and $s^f(u_t)$ be scores of $u_t \in U$ in the weighted matrices $\wedge^t, \wedge^i$ and $\wedge^f$. Then,
 decision score of $u_t \in U$, denoted by $ds(u_t)$, is computed  as follow:

$$
ds(u_t)=s^t(u_t)-s^i(u_t)-s^f(u_t)
$$
\end{defn}
Now, we construct a possibility neutrosophic soft set  decision making method by
the following algorithm;
\section*{Algorithm}

\emph{\textbf{Step 1:}} Input the possibility neutrosophic soft set $f_{\mu}$,\\

\emph{\textbf{Step 2:}} Construct the matrix $\wedge$-product\\

\emph{\textbf{Step 3:}} Construct the matrices $f^t_{\mu},f^i_{\mu}$ and $f^f_{\mu}$\\

\emph{\textbf{Step 4:}} Construct the weighted  matrices $\wedge^t,\wedge^i$ and $\wedge^f$, \\

\emph{\textbf{Step 5:}} Compute score of $u_t\in U$,  for each of the weighted  matrices,   \\

\emph{\textbf{Step 6:}} Compute decision score, for all $u_t\in U$,\\

\emph{\textbf{Step 7:}} The optimal decision is to select $u_t=max{ds(u_i)}$.\\

\begin{exmp} Assume that $U=\{u_1,u_2,u_3\}$ is a set of houses and
$E=\{e_1,e_2,e_3\}=\{cheap, \, large,\, moderate\}$ is a set of
parameters which is attractiveness of houses. Suppose that Mr.X want
to buy one the most suitable house according to to himself depending
on three of the  parameters only. \\

\emph{\textbf{Step 1:}}Based on the choice parameters of
$Mr.X$, let there be two observations $f_{\mu}$ and $g_{\nu}$ by two
experts defined as follows:

 \small $f_{\mu}=\left\{ \begin{array}{c}
f_{\mu}(e_1)=\Big\{\Big(\frac{u_1}{(0.5,0.3,0.7)},0.6\Big),\Big(\frac{u_2}{(0.6,0.2,0.5)},0.2\Big), \Big(\frac{u_3}{(0.7,0.6,0.5)},0.4\Big)\Big\} \\
f_{\mu}(e_2)=\Big\{\Big(\frac{u_1}{(0.35,0.2,0.6)},0.4\Big),\Big(\frac{u_2}{(0.7,0.8,0.3)},0.5\Big), \Big(\frac{u_3}{(0.2,0.4,0.4)},0.6\Big)\Big\} \\
f_{\mu}(e_3)=\Big\{\Big(\frac{u_1}{(0.7,0.2,0.5)},0.5\Big), \Big(\frac{u_2}{(0.4,0.5,0.2)}, 0.3\Big),\Big(\frac{u_3}{(0.5,0.3,0.6)},0.2\Big) \Big\}\\
\end{array}\right\}$\\
\\

$g_{\nu}=\left\{\begin{array}{c}
g_{\nu}(e_1)=\Big\{\Big(\frac{u_1}{(0.3,0.4,0.5)},0.2\Big),\Big(\frac{u_2}{(0.7,0.3,0.4)},0.5\Big), \Big(\frac{u_3}{(0.4,0.5,0.2)},0.3\Big)\Big\} \\
g_{\nu}(e_2)=\Big\{\Big(\frac{u_1}{(0.4,0.6,0.2)},0.3\Big),\Big(\frac{u_2}{(0.2,0.5,0.3)},0.7\Big), \Big(\frac{u_3}{(0.4,0.6,0.2)},0.8\Big)\Big\} \\
g_{\nu}(e_3)=\Big\{\Big(\frac{u_1}{(0.2,0.1,0.6)},0.7\Big), \Big(\frac{u_2}{(0.8,0.4,0.5)}, 0.4\Big),\Big(\frac{u_3}{(0.6,0.4,0.3)},0.4\Big) \Big\}\\
\end{array}\right\}$\\

\emph{\textbf{Step 2:}}
\normalsize Let us consider possibility neutrosophic soft set
$\wedge$-product which is the mapping $\wedge:E\times E\to
\mathcal{N(U)}\times I^U$ given as follows:

$$\left(
\begin{array}{c|ccc}
 \wedge &  u_1,\mu &
u_2,\mu&  u_3,\mu \\
    \hline e_{11}  &
(\langle 0.3,0.4,0.7 \rangle, 0.2)& (\langle 0.6,0.3,0.5 \rangle, 0.2)& (\langle 0.4,0.6,0.5 \rangle, 0.3) \\
    e_{12} &
(\langle 0.4,0.6,0.7 \rangle, 0.3)& (\langle 0.2,0.5,0.5 \rangle, 0.2)& (\langle 0.4,0.6,0.5 \rangle, 0.4) \\
    e_{13}  &
(\langle 0.2,0.3,0.7 \rangle, 0.6)& (\langle 0.6,0.4,0.5 \rangle, 0.2)& (\langle 0.6,0.6,0.5 \rangle, 0.4) \\
e_{21}  &
(\langle 0.3,0.4,0.6 \rangle, 0.2)& (\langle 0.7,0.8,0.4 \rangle, 0.5)& (\langle 0.2,0.5,0.5 \rangle, 0.3) \\
  e_{22}  &
(\langle 0.35,0.6,0.6 \rangle, 0.3)& (\langle 0.2,0.8,0.3 \rangle, 0.5)& (\langle 0.2,0.6,0.5 \rangle, 0.6) \\
   e_{23}  &
(\langle 0.2,0.2,0.6 \rangle, 0.4)& (\langle 0.7,0.8,0.5 \rangle, 0.4)& (\langle 0.2,0.4,0.5 \rangle, 0.4) \\
    e_{31} &
(\langle 0.3,0.4,0.5 \rangle, 0.2)& (\langle 0.4,0.5,0.4 \rangle, 0.3)& (\langle 0.4,0.5,0.6 \rangle, 0.2) \\
    e_{32}  &
(\langle 0.4,0.6,0.5 \rangle, 0.3)& (\langle 0.2,0.5,0.3 \rangle, 0.3)& (\langle 0.4,0.6,0.6 \rangle, 0.2) \\
e_{33}  &
(\langle 0.2,0.2,0.6 \rangle, 0.5)& (\langle 0.4,0.5,0.5 \rangle, 0.3)& (\langle 0.5,0.4,0.6 \rangle, 0.2) \\
   \end{array}\right)
   $$
\begin{center}
\footnotesize{ Matrix representation of  $\wedge$-product}
\end{center}

\emph{\textbf{Step 3:}} We construct matrices $f^t_{\mu},f^i_{\mu}$ and $f^f_{\mu}$  as follows:
$$\left(
\begin{array}{c|ccc}
 \wedge &  u_1,\mu &
u_2,\mu&  u_3,\mu \\
    \hline e_{11}  &
(0.3, 0.2)& (0.6, 0.2)& (0.4, 0.3) \\
    e_{12} &
(0.4, 0.3)& (0.2, 0.2)& (0.4, 0.4) \\
    e_{13}  &
(0.2, 0.6)& (0.6, 0.2)& (0.6, 0.4) \\
e_{21}  &
(0.3, 0.2)& (0.7, 0.5)& (0.2, 0.3) \\
  e_{22}  &
(0.35, 0.3)& (0.2, 0.5)& (0.2, 0.6) \\
   e_{23}  &
(0.2, 0.4)& (0.7, 0.4)& (0.2, 0.4) \\
    e_{31} &
(0.3, 0.2)& (0.4, 0.3)& (0.4, 0.2) \\
    e_{32}  &
(0.4, 0.3)& (0.2, 0.3)& (0.4, 0.2) \\
e_{33}  &
(0.2, 0.5)& (0.4, 0.3)& (0.5, 0.2) \\
   \end{array}\right)
   $$
\begin{center}
\footnotesize{ Matrix $f_{\mu}^t$ of $\wedge$-product }
\end{center}

$$\left(
\begin{array}{c|ccc}
 \wedge &  u_1,\mu &
u_2,\mu&  u_3,\mu \\
    \hline e_{11}  &
(0.4, 0.2)& (0.3, 0.2)& (0.6, 0.3) \\
    e_{12} &
(0.6, 0.3)& (0.5, 0.2)& (0.6, 0.4) \\
    e_{13}  &
(0.3, 0.6)& (0.4, 0.2)& (0.6, 0.4) \\
e_{21}  &
(0.4, 0.2)& (0.8, 0.5)& (0.5, 0.3) \\
  e_{22}  &
(0.6, 0.3)& (0.8, 0.5)& (0.6, 0.6) \\
   e_{23}  &
(0.2, 0.4)& (0.8, 0.4)& (0.4, 0.4) \\
    e_{31} &
(0.4, 0.2)& (0.5, 0.3)& (0.5, 0.2) \\
    e_{32}  &
(0.6, 0.3)& (0.5, 0.3)& (0.6, 0.2) \\
e_{33}  &
(0.2, 0.5)& (0.5, 0.3)& (0.4, 0.2) \\
   \end{array}\right)
   $$
\begin{center}
\footnotesize{Matrix $f_{\mu}^i$ of $\wedge$-product }
\end{center}

$$\left(
\begin{array}{c|ccc}
 \wedge &  u_1,\mu &
u_2,\mu&  u_3,\mu \\
    \hline e_{11}  &
(0.7, 0.2)& (0.5, 0.2)& (0.5, 0.3) \\
    e_{12} &
(0.7, 0.3)& (0.5, 0.2)& (0.5, 0.4) \\
    e_{13}  &
(0.7, 0.6)& (0.5, 0.2)& (0.5, 0.4) \\
e_{21}  &
(0.6, 0.2)& (0.4, 0.5)& (0.5, 0.3) \\
  e_{22}  &
(0.6, 0.3)& (0.3, 0.5)& (0.5, 0.6) \\
   e_{23}  &
(0.6, 0.4)& (0.5, 0.4)& (0.5, 0.4) \\
    e_{31} &
(0.5, 0.2)& (0.4, 0.3)& (0.6, 0.2) \\
    e_{32}  &
(0.5, 0.3)& (0.3, 0.3)& (0.6, 0.2) \\
e_{33}  &
(0.6, 0.5)& (0.5, 0.3)& (0.6, 0.2) \\
   \end{array}\right)
   $$
\begin{center}
\footnotesize{ Matrix $f_{\mu}^f$ of $\wedge$-product}
\end{center}

\emph{\textbf{Step 4:}}
We  obtain  weighted  matrices $\wedge^t,\wedge^i$ and $\wedge^f$ using Definition \ref{weighedmatrices}  as follows:

\footnotesize
$$\left(
\begin{array}{c|ccc}
 \wedge^t &  u_1 &
u_2&  u_3\\
    \hline e_{11}  &
0.44 & \underline{0.64} & 0.58 \\
    e_{12} &
0.58 & 0.36 & \underline{0.64} \\
    e_{13}  &
0.68 & 0.68 & \underline{0.76 }\\
e_{21}  &
0.44 & \underline{0.85} & 0.44 \\
  e_{22}  &
0.55 & 0.60 & \underline{0.68} \\
   e_{23}  &
0.52 & \underline{0.82} & 0.48 \\
    e_{31} &
0.44 & \underline{0.58} & 0.52 \\
    e_{32}  &
\underline{0.58} & 0.44 & 0.52 \\
e_{33}  &
\underline{0.60} & 0.58 & \underline{0.60} \\
   \end{array}\right),
   \quad
   \left(
   \begin{array}{c|ccc}
 \wedge^i &  u_1 &
u_2 &  u_3 \\
    \hline e_{11}  &
0.08 & \underline{0.16} & 0.18 \\
    e_{12} &
0.18 & 0.10 & \underline{0.24} \\
    e_{13}  &
0.18 & 0.08 & \underline{0.24} \\
e_{21}  &
0.08 & \underline{0.40} & 0.15 \\
  e_{22}  &
0.18 & \underline{0.40} & 0.36 \\
   e_{23}  &
0.08 & \underline{0.32} & 0.16 \\
    e_{31} &
0.08 & \underline{0.15} & 0.10 \\
    e_{32}  &
\underline{0.18} & 0.15 & 0.12 \\
e_{33}  &
0.10 & \underline{0.15} & 0.08\\
   \end{array}\right),
   \quad
    \left(
   \begin{array}{c|ccc}
 \wedge^f &  u_1 &
u_2 &  u_3 \\
    \hline e_{11}  &
0.14 & 0.10 & \underline{0.15} \\
    e_{12} &
\underline{0.21} & 0.10 & 0.20 \\
    e_{13}  &
\underline{0.42} & 0.10 & 0.20 \\
e_{21}  &
0.12 & \underline{0.20} & 0.15 \\
  e_{22}  &
0.18 & 0.15 & \underline{0.30} \\
   e_{23}  &
\underline{0.24} & 0.20 & \underline{0.20} \\
    e_{31} &
0.10 & \underline{0.12} & \underline{0.12} \\
    e_{32}  &
\underline{0.15} & 0.09 & 0.12 \\
e_{33}  &
\underline{0.30} & 0.15 & 0.12 \\
   \end{array}\right)
   $$
\begin{center}
\footnotesize{ Weighed matrices of $f_{\mu}^t,f_{\mu}^i$ and
$f_{\mu}^f$ from left to right, respectively. }
\end{center}
\normalsize

\emph{\textbf{Step 5:}} For all $u\in U$, we find scores using Definition \ref{score} as follow:

$$s^t(u_1)=1,18, \quad s^t(u_2)=2,89,\quad s^t(u_3)=2,68$$
$$s^i(u_1)=0,18 \quad s^i(u_2)=1,42 \quad s^i(u_3)=0,66$$
$$s^f(u_1)=1,32 \quad s^f(u_2)=0,32 \quad s^f(u_3)=0,57$$

\normalsize
\emph{\textbf{Step 5:}} For all $u\in U$, we find scores using Definition \ref{score} as follows:
\begin{eqnarray*}
ds(u_1)&=&1,18-0,18-1,32=-0,32\\
ds(u_2)&=&2,89-1,42-0,32=0,90\\
ds(u_3)&=&2,68-0,66-0,57=1,45\\
\end{eqnarray*}

\emph{\textbf{Step 5:}}
Then the optimal selection for $Mr.X$ is $u_3.$

\end{exmp}

\section{Similarity measure of possibility neutrosophic soft sets}
In this section, we introduce a measure of similarity between two
$PNS$-sets.

\begin{defn}\label{defn-similarity} Similarity between two $PNS$-sets $f_{\mu}$ and
$g_{\nu}$, denoted by $S(f_{\mu},g_{\nu})$, is defined as follows:

$$
S(f_{\mu},g_{\nu})=M(f(e),g(e)).M(\mu(e),\nu(e))
$$
such that
$$
M(f(e),g(e))=\frac{1}{n}M_i(f(e),g(e)),
M(\mu,\nu)=\frac{1}{n}\sum_{i=1}^nM(\mu(e_i),\nu(e_i)),
$$
where
$$
M_i(f(e),g(e))=1-\frac{1}{\sqrt[p]{n}}\sqrt[p]{\sum_{i=1}^n(\phi_{f_{\mu}(e_i)}(u_j)-\phi_{g_{\nu}(e_i)}(u_j))^p},
1\leq p\leq \infty
$$
\end{defn}
such that and
$$
\phi_{f_{\mu}(e_i)}(u_j)=\frac{f^t_{ij}(e_i)+f^i_{ij}(e_i)+f^f_{ij}(e_i)}{3},
\,\,
\phi_{g_{\mu}(e_i)}(u_j)=\frac{g^t_{ij}(e_i)+g^i_{ij}(e_i)+g^f_{ij}(e_i)}{3},
$$
$$
M(\mu(e_i),\nu(e_i))=1-\frac{\sum_{j=1}^n|\mu_{ij}(e_i)-\nu_{ij}(e_i)|}{\sum_{j=1}^n|\mu_{ij}(e_i)+\nu_{ij}(e_i)|}
$$

\begin{defn} Let $f_{\mu}$ and $g_{\nu}$ be two PNS-sets over $U$.
We say that $f_{\mu}$ and $g_{\nu}$ are significantly similar if
$S(f_{\mu},g_{\nu})\geq\frac{1}{2}$
\end{defn}

\begin{prop} Let $f_{\mu},g_{\nu}\in \mathcal{PN}(U,E)$. Then,
\begin{enumerate}[i.]
\item $S(f_{\mu},g_{\nu})=S(g_{\mu},f_{\nu})$
\item $0\leq S(f_{\mu},g_{\nu})\leq 1 $
\item $f_{\mu}=g_{\nu}\Rightarrow S(f_{\mu},g_{\nu})=1$
\end{enumerate}
\end{prop}
\begin{pf}The proof is straightforward and follows from Definition
\ref{defn-similarity}.
\end{pf}

\begin{exmp} Let us consider PNS-sets $f_{\mu}$ and $g_{\nu}$ in  Example \ref{pnss-ex}
given as follows:\\

 $\left\{
\begin{array}{c}
f_{\mu}(e_1)=\Big\{\Big(\frac{u_1}{(0.5,0.2,0.6)},
0.8\Big),\Big(\frac{u_2}{(0.7,0.3,0.5)},
0.4\Big),\Big(\frac{u_3}{(0.4,0.5,0.8)},
0.7\Big)\Big\} \\
f_{\mu}(e_2)=\Big\{\Big(\frac{u_1}{(0.8,0.4,0.5)},
0.6\Big),\Big(\frac{u_2}{(0.5,0.7,0.2)},
0.8\Big),\Big(\frac{u_3}{(0.7,0.3,0.9)},
0.4\Big)\Big\}\\
f_{\mu}(e_3)=\Big\{\Big(\frac{u_1}{(0.6,0.7,0.5)},
0.2\Big),\Big(\frac{u_2}{(0.5,0.3,0.7)},
0.6\Big),\Big(\frac{u_3}{(0.6,0.5,0.4)}, 0.5\Big)\Big\}
\end{array}\right\}$\\
and\\

 $\left\{\begin{array}{c}
g_{\nu}(e_1)=\Big\{\Big(\frac{u_1}{(0.6,0.3,0.8)},
0.4\Big),\Big(\frac{u_2}{(0.6,0.5,0.5)},
0.7),\Big(\frac{u_3}{(0.2,0.6,0.4)},
0.8\Big)\Big\} \\
g_{\nu}(e_2)=\Big\{\Big(\frac{u_1}{(0.5,0.4,0.3)},
0.3\Big),\Big(\frac{u_2}{(0.4,0.6,0.5)},
0.6\Big),\Big(\frac{u_3}{(0.7,0.2,0.5)},
0.8\Big)\Big\}\\
g_{\nu}(e_3)=\Big\{\Big(\frac{u_1}{(0.7,0.5,0.3)},
0.8\Big),\Big(\frac{u_2}{(0.4,0.4,0.6)},
0.5\Big),\Big(\frac{u_3}{(0.8,0.5,0.3)}, 0.6\Big)\Big\}
\end{array}\right\}$
\end{exmp}
then,

\begin{eqnarray*}
M(\mu(e_1),\nu(e_1))&=&1-\frac{\sum_{j=1}^3|\mu_{1j}(e_1)-\nu_{1j}(e_1)|}{\sum_{j=1}^3|\mu_{1j}(e_1)+\nu_{1j}(e_1)|}\\
&=&1-\frac{|0.8-0.4|+|0.4-0.7|+|0.7-0.8|}{|0.8+0.4|+|0.4+0.7|+|0.7+0.8|}=0.79
\end{eqnarray*}
Similarly we get $M(\mu(e_2),\nu(e_2))=0.74$ and
$M(\mu(e_3),\nu(e_3))=0.75$, then

$$
M(\mu,\nu)=\frac{1}{3}(0.79+0.75+0.74)=0.76
$$

\begin{eqnarray*}
M_1(f(e),g(e))&=&1-\frac{1}{\sqrt[p]{n}}\sqrt[p]{\sum_{i=1}^n(\phi_{f_{\mu}(e_i)}(u_j)-\phi_{g_{\nu}(e_i)}(u_j))^p}\\
&=&1-\frac{1}{\sqrt{3}}\sqrt{(0.43-0.57)^2+(0,50-0.53)^2+(0,57-0,40)^2}=0.73\\
M_2(f(e),g(e))&=&0.86\\
M_3(f(e),g(e))&=&0.94
\end{eqnarray*}
$$
M(f(e),g(e))=\frac{1}{3}(0.73+0.86+0.94)=0.84
$$
and
$$
S(f_{\mu},g_{\nu})=0.84\times0.76=0.64
$$

\section{Decision-making method based on the similarity measure}
 In this section, we give a decision making problem involving
 possibility neutrosophic soft sets by means of the similarity
 measure between the possibility neutrosophic soft sets.

Let our universal set contain only two elements "yes" and "no", that
is $U={y,n}$. Assume that $P=\{p_1,p_2,p_3,p_4,p_5\}$ are five
candidates who fill in a form in order to apply formally for the
position. There is a decision maker committee. They want to
interview  the candidates by model possibility neutrosophic soft set
determined by committee. So they want to test similarity of each of
candidate to model possibility neutrosophic soft set.

Let $E=\{e_1,e_2,e_3,e_4,e_5,e_6,e_7\}$ be the set of parameters,
where   $e_1$=experience, $e_2$=computer knowledge, $e_3$=training,
$e_4$=young age, $e_5$=higher education, $e_6$=marriage status and
$e_7$=good health.

Our model possibility neutrosophic soft set determined by committee
for suitable candidates properties $f_{\mu}$ is given in Table 1.
\footnotesize
$$
\begin{tabular}{|c|c|c|c|c|}
  \hline
 \(f_{\mu}\) &  \(e_1, \mu\) &
\(e_2, \mu\)&  \(e_3, \mu\) & \(e_4, \mu\) \\
    \hline \(y\)  &
\((\langle 1,0,0 \rangle, 1 )\)& \((\langle 1,0,0 \rangle, 1 )\)&
\((\langle 0,1,1 \rangle, 1 )\)&
\((\langle 0,1,1 \rangle, 1 )\)\\
\hline \(n\) & \((\langle 0,1,1 \rangle, 1 )\)& \((\langle 1,0,0
\rangle, 1 )\)&
 \((\langle 0,1,1 \rangle, 1 )\) & \((\langle 1,0,0 \rangle, 1 )\)\\
\hline
    \end{tabular}
$$
$$
\begin{tabular}{|c|c|c|c|}
\hline  \(f_{\mu}\) & \(e_5, \mu\) & \(e_6, \mu\) &  \(e_7, \mu\) \\
\hline \(y\)  &  \((\langle 1,0,0 \rangle, 1 )\)& \((\langle 0,1,1
\rangle, 1 )\) &
 \((\langle 1,0,0 \rangle, 1 )\)\\
\hline \(n\) & \((\langle 0,1,1 \rangle, 1 )\)& \((\langle 1,0,0
\rangle, 1 )\)&
 \((\langle 0,1,1 \rangle, 1 )\)  \\
\hline
\end{tabular}
$$
\begin{center}
\footnotesize{\emph{\emph{Table 1}: The tabular representation of
model possibility neutrosophic soft set}}
\end{center}

\footnotesize
$$
\begin{tabular}{|c|c|c|c|c|}
  \hline
   \(g_{\nu}\) &  \(e_1, \nu\) &
\(e_2, \nu\)&  \(e_3, \nu\) & \(e_4, \nu\) \\
    \hline \(y\)  &
\((\langle 0.7,0.2,0.5 \rangle, 0.4 )\)& \((\langle 0.5,0.4,0.6
\rangle, 0.2 )\)& \((\langle 0.2,0.3,0.4 \rangle, 0.5 )\)&
\((\langle 0.8,0.4,0.6 \rangle, 0.3 )\) \\
\hline \(n\) & \((\langle 0.3,0.7,0.1 \rangle, 0.3 )\)& \((\langle
0.7,0.3,0.5 \rangle, 0.4 )\)& \((\langle 0.6,0.5,0.3 \rangle, 0.2
)\)& \((\langle 0.2,0.1,0.5 \rangle, 0.4 )\) \\
\hline
    \end{tabular}
$$
$$
\begin{tabular}{|c|c|c|c|}
\hline  \(g_{\nu}\) & \(e_5, \nu\) & \(e_6, \nu\) &  \(e_7, \nu\) \\
\hline \(y\)  &  \((\langle 0.2,0.4,0.3 \rangle, 0.5 )\)& \((\langle 0,1,1 \rangle, 0.3 )\) & \((\langle 0.1,0.4,0.7 \rangle, 0.2 )\)\\
\hline \(n\) & \((\langle 0.1,0.5,0.2 \rangle, 0.6 )\) & \((\langle
1,0,0 \rangle, 0.5 )\)&
 \((\langle 0.3,0.5,0.1 \rangle, 0.4)\) \\
\hline
\end{tabular}
$$
\begin{center}
\footnotesize{\emph{\emph{Table 2}: The tabular representation of
possibility neutrosophic soft set for $p_1$}}
\end{center}

$$
\begin{tabular}{|c|c|c|c|c|}
  \hline
   \(h_{\delta}\) &  \(e_1, \delta\) &
\(e_2, \delta\)&  \(e_3, \delta\) & \(e_4, \delta\) \\
    \hline \(y\)  &
\((\langle 0.8,0.2,0.1 \rangle, 0.3 )\)& \((\langle 0.4,0.2,0.6
\rangle, 0.1 )\)& \((\langle 0.7,0.2,0.4 \rangle, 0.2 )\)&
\((\langle 0.3,0.2,0.7 \rangle, 0.6 )\) \\
\hline \(n\) & \((\langle 0.2,0.4,0.3 \rangle, 0.5 )\)& \((\langle
0.6,0.3,0.2 \rangle, 0.3 )\)& \((\langle 0.4,0.3,0.2 \rangle, 0.1
)\)& \((\langle 0.8,0.1,0.3 \rangle, 0.3 )\) \\
\hline
    \end{tabular}
$$
$$
\begin{tabular}{|c|c|c|c|}
\hline  \(h_{\delta}\) & \(e_5, \delta\) & \(e_6, \delta\) &  \(e_7, \delta \) \\
\hline \(y\)  &  \((\langle 0.5,0.2,0.4 \rangle, 0.5 )\)& \((\langle 0,1,1 \rangle, 0.5 )\) & \((\langle 0.3,0.2,0.5 \rangle, 0.4 )\)\\
\hline \(n\) & \((\langle 0.4,0.5,0.6 \rangle, 0.2 )\) & \((\langle
1,0,0 \rangle, 0.2 )\)&  \((\langle 0.7,0.3, 0.4 \rangle, 0.2)\) \\
\hline
\end{tabular}
$$
\begin{center}
\footnotesize{\emph{\emph{Table 3}: The tabular representation of
possibility neutrosophic soft set for $p_2$}}
\end{center}

$$
\begin{tabular}{|c|c|c|c|c|}
  \hline
   \(r_{\theta}\) &  \(e_1, \theta\) &
\(e_2, \theta\)&  \(e_3, \theta\) & \(e_4, \theta\) \\
    \hline \(y\)  &
\((\langle 0.3,0.2,0.5 \rangle, 0.4 )\)& \((\langle 0.7,0.1,0.5
\rangle, 0.6 )\)& \((\langle 0.6,0.5,0.3 \rangle, 0.2 )\)&
\((\langle 0.3,0.1,0.4 \rangle, 0.5 )\) \\
\hline \(n\) & \((\langle 0.1,0.7,0.6 \rangle, 0.3 )\)& \((\langle
0.4,0.2,0.3 \rangle, 0.7 )\)& \((\langle 0.7,0.4,0.3 \rangle, 0.5
)\)& \((\langle 0.7,0.1,0.2 \rangle, 0.1 )\) \\
\hline
    \end{tabular}
$$
$$
\begin{tabular}{|c|c|c|c|}
\hline  \(r_{\theta}\) & \(e_5, \theta\) & \(e_6, \theta\) &  \(e_7, \theta\) \\
\hline \(y\)  &  \((\langle 0.6,0.4,0.3 \rangle, 0.2 )\)& \((\langle 0,1,1 \rangle, 0.3 )\) & \((\langle 0.9,0.1,0.1 \rangle, 0.5 )\)\\
\hline \(n\) & \((\langle 0.4,0.5,0.9 \rangle, 0.1 )\) & \((\langle
1,0,0 \rangle, 0.3 )\)&  \((\langle 0.2,0.1, 0.7 \rangle, 0.6)\) \\
\hline
\end{tabular}
$$
\begin{center}
\footnotesize{\emph{\emph{Table 4}: The tabular representation of
possibility neutrosophic soft set for $p_3$}}
\end{center}

$$
\begin{tabular}{|c|c|c|c|c|}
  \hline
   \(s_{\alpha}\) &  \(e_1, \alpha\) &
\(e_2, \alpha\)&  \(e_3, \alpha\) & \(e_4, \alpha\) \\
    \hline \(y\)  &
\((\langle 0.2,0.1,0.4 \rangle, 0.5 )\)& \((\langle 0.7,0.5,0.4
\rangle, 0.8 )\)& \((\langle 0.8,0.1,0.2 \rangle, 0.4 )\)&
\((\langle 0.5,0.4,0.5 \rangle, 0.4 )\) \\
\hline \(n\) & \((\langle 0.6,0.5,0.1 \rangle, 0.1 )\)& \((\langle
0.3,0.7,0.2 \rangle, 0.2 )\)& \((\langle 0.7,0.5,0.1 \rangle, 0.7
)\)& \((\langle 0.1,0.3,0.7 \rangle, 0.5 )\) \\
\hline
    \end{tabular}
$$
$$
\begin{tabular}{|c|c|c|c|}
\hline  \(s_{\alpha}\) & \(e_5, \alpha\) & \(e_6, \alpha\) &  \(e_7, \alpha\) \\
\hline \(y\)  &  \((\langle 0.3,0.2,0.5 \rangle, 0.8 )\)& \((\langle 1,0,0 \rangle, 0.7 )\) & \((\langle 0.1,0.8,0.9 \rangle, 0.7 )\)\\
\hline \(n\) & \((\langle 0.2,0.1,0.5 \rangle, 0.3 )\) & \((\langle
0,1,1 \rangle, 0.2 )\)&  \((\langle 0.5,0.1, 0.4 \rangle, 0.1)\) \\
\hline
\end{tabular}
$$
\begin{center}
\footnotesize{\emph{\emph{Table 5}: The tabular representation of
possibility neutrosophic soft set for $p_4$}}
\end{center}

$$
\begin{tabular}{|c|c|c|c|c|}
  \hline
   \(m_{\gamma}\) &  \(e_1, \gamma\) &
\(e_2, \gamma\)&  \(e_3, \gamma\) & \(e_4, \gamma\) \\
    \hline \(y\)  &
\((\langle 0.1,0.2,0.1 \rangle, 0.3 )\)& \((\langle 0.2,0.3,0.5
\rangle, 0.8 )\)& \((\langle 0.4,0.1,0.3 \rangle, 0.9 )\)&
\((\langle 0.7,0.3,0.2 \rangle, 0.3 )\) \\
\hline \(n\) & \((\langle 0.4,0.5,0.3 \rangle, 0.2 )\)& \((\langle
0.7,0.6,0.1 \rangle, 0.3 )\)& \((\langle 0.2,0.3,0.4 \rangle, 0.5
)\)& \((\langle 0.5,0.2,0.3 \rangle, 0.6 )\) \\
\hline
    \end{tabular}
$$
$$
\begin{tabular}{|c|c|c|c|}
\hline  \(m_{\gamma}\) & \(e_5, \gamma\) & \(e_6, \gamma\) &  \(e_7, \gamma\) \\
\hline \(y\)  &  \((\langle 0.4,0.2,0.8 \rangle, 0.1 )\)& \((\langle 1,0,0 \rangle, 0.5 )\) & \((\langle 0.3,0.2,0.1 \rangle, 0.7 )\)\\
\hline \(n\) & \((\langle 0.5,0.4,0.7 \rangle, 0.2 )\) & \((\langle
0,1,1 \rangle, 0.5 )\)&  \((\langle 0.3,0.2, 0.1 \rangle, 0.9)\) \\
\hline
\end{tabular}
$$
\begin{center}
\footnotesize{\emph{\emph{Table 6}: The tabular representation of
possibility neutrosophic soft set for $p_5$}}
\end{center}
\normalsize Now we find the similarity between the model possibility
neutrosophic soft set  and  possibility neutrosophic soft set of
each person as follow

 $S(f_{\mu},g_{\nu})\cong0,49< \frac{1}{2}$,
$S(f_{\mu},h_{\delta})\cong0,47< \frac{1}{2}$,
$S(f_{\mu},r_{\theta})\cong0,51> \frac{1}{2} $,
$S(f_{\mu},s_{\alpha})\cong0,54> \frac{1}{2}$,
$S(f_{\mu},m_{\gamma})\cong0,57> \frac{1}{2}$,\\

Consequently, $p_5$ is should  be selected by the committee.

\section{Conclusion}
In this paper we have introduced  the concept of possibility
neutrosophic soft set and studied some of the related  properties.
Applications of this theory have been given to solve a decision
making problem. We also  presented a new method  to find out the
similarity measure of two possibility neutrosophic soft sets and we
applied to a decision making problem. In future, these seem to have
 natural applications  and algebraic structure.


\begin{thebibliography}{99}
\bibitem{deli-14ivnss} I. Deli,Interval-valued neutrosophic soft sets ant its decision making, arxiv:1402.3130
\end{thebibliography}
\end{document}